\documentclass[12pt]{article}
\usepackage{amssymb,tabularx,multirow,amsmath, epsfig, threeparttable, amstext}
\usepackage{amssymb}		
\usepackage{amsthm}			
\usepackage{tocloft}		
\usepackage{float}			
\usepackage{sectsty}		
\usepackage{graphicx}		
\usepackage{setspace}		
\usepackage{setspace}
\usepackage{verbatim}
\usepackage{times}
\usepackage{media9} 
\usepackage{framed}
\usepackage{caption}
\usepackage{subcaption}
\usepackage{algorithm2e}
\usepackage{bbm}
\usepackage{varwidth}
\usepackage{cases}

\usepackage[natbibapa]{apacite}

\usepackage[usenames, dvipsnames]{color}

\SetKwInOut{Input}{Input}
\SetKwInOut{Output}{Output}
\SetKwInOut{Data}{Data}
\makeatletter
\newcommand*\bigcdot{\mathpalette\bigcdot@{.5}}
\newcommand*\bigcdot@[2]{\mathbin{\vcenter{\hbox{\scalebox{#2}{$\m@th#1\bullet$}}}}}
\makeatother

\newcommand{\R}{\mathbb{R}}

\newcommand{\mbf}[1]{\mathbf{#1}}
\newcommand{\mbv}[1]{\mbox{\boldmath$#1$\unboldmath}}

\def\bh{\mathbf{h}}

\def\br{\mathbf{r}}
\def\bs{\mathbf{s}}

\def\bv{\mathbf{v}}

\def\bx{\mathbf{x}}

\def\bI{\mathbf{I}}

\def\bM{\mathbf{M}}

\def\bV{\mathbf{V}}
\def\bU{\mathbf{U}}
\def\bW{\mathbf{W}}

\def\bZ{\mathbf{Z}}

\newcommand{\bfalpha}{\mbox{\boldmath $\alpha$}}

\newcommand{\bfPhi}{\mbox{\boldmath $\Phi$}}

\usepackage{color}
\usepackage{bm}
\usepackage{multirow}
\usepackage{amsmath}
\usepackage{amsfonts}
\usepackage{setspace}
\usepackage{lineno}
\usepackage{fancyhdr}

\fancypagestyle{allpage}{
  \fancyhf{}
  \fancyfoot[C]{\ifnum\value{page}<0\relax\else\thepage\fi}
  \renewcommand{\headrulewidth}{0pt}
  \fancyhead[R]{P. L. MCDERMOTT AND C. K. WIKLE}
}

\begin{document}
\bibliographystyle{apacite}

\doublespacing
\begin{titlepage}

\fancypagestyle{firstpage}{
  \fancyhf{}
   \fancyfoot[C]{\thepage}
  \renewcommand{\headrulewidth}{0pt}
}

\title{ \vspace{-8ex}  Deep Echo State Networks with Uncertainty Quantification for Spatio-Temporal Forecasting}
\date{\vspace{-8.5ex}}
\author{Patrick L. McDermott\thanks{Correspondence to: P.L. McDermott, Jupiter Intelligence
Boulder, CO 80302 U.S.A., E-mail: plmyt7@gmail.com} \footnotemark[2] ,\,\, Christopher K. Wikle\thanks{Department of Statistics, University of Missouri, Columbia, MO, 65211, U.S.A.}\vspace{.55cm} }

 \maketitle
 \thispagestyle{firstpage}
\begin{abstract}
Long-lead forecasting for spatio-temporal systems can often entail complex nonlinear dynamics that are difficult to specify {\it a priori}. Current statistical methodologies for modeling these processes are often highly parameterized and thus, challenging to implement from a computational perspective. One potential parsimonious solution to this problem is a method from the dynamical systems and engineering literature referred to as an echo state network (ESN). ESN models use so-called {\it reservoir computing} to efficiently compute recurrent neural network (RNN) forecasts.   Moreover, so-called ``deep" models have recently been shown to be successful at predicting high-dimensional complex nonlinear processes, particularly those with multiple spatial and temporal scales of variability (such as we often find in spatio-temporal environmental data).  Here we introduce a deep ensemble ESN (D-EESN) model.  We present two versions of this model for spatio-temporal processes that both produce forecasts and associated measures of uncertainty.  The first approach utilizes a bootstrap ensemble framework and the second is developed within a hierarchical Bayesian framework (BD-EESN).  This more general hierarchical Bayesian framework naturally accommodates non-Gaussian data types and multiple levels of uncertainties. The methodology is first applied to a data set simulated from a novel non-Gaussian multiscale Lorenz-96 dynamical system simulation model and then to a long-lead  United States (U.S.) soil moisture forecasting application.
 \end{abstract}
\baselineskip=22pt
\end{titlepage}
\setcounter{page}{2}
\pagestyle{allpage}
\section{Introduction}

Spatio-temporal data are ubiquitous in engineering and the sciences, and their study is important for understanding and predicting a wide variety of processes.   One of the chief difficulties in modeling spatial processes that change with time is the complexity of the dependence structures that must describe how such a process varies, and the presence of high-dimensional complex datasets and large prediction domains.  It is particularly challenging to specify parameterizations for nonlinear dynamical spatio-temporal models that are simultaneously useful scientifically (e.g., long-lead forecasting as discussed below) and efficient computationally. Statisticians have developed some ``deep'' mechanistically-motivated hierarchical models that can accommodate process complexity as well as the uncertainties in predictions and inference, typically within a hierarchical Bayesian paradigm \citep[see the overview in][]{CandW2011}.  However, these models can be computationally expensive, require prior information and/or a significant amount of data to fit, and are typically application specific.   On the other hand, the science, engineering, and machine learning communities have developed alternative approaches for nonlinear spatio-temporal modeling, particularly in the neural network context (e.g., recurrent neural networks, RNNs).  These approaches can be very flexible but, again, are computationally expensive, require a lot of training data, and/or ``pre-training.''  In addition, these approaches often do not provide formal measures of uncertainty quantification.  There are, however, parsimonious approaches to RNNs in the engineering literature, such as the echo state network (ESN), although the standard implementation of this approach does not include spatio-temporal dependencies, deep learning, or formal uncertainty quantification.  Here, we present a hierarchical deep statistical implementation of an ESN model for spatio-temporal processes with the goal of long-lead forecasting with uncertainty quantification.

The methodology presented here is motivated by the problem of long-lead forecasting of environmental processes. The atmosphere is a chaotic dynamical system.  Because of that, skillful weather forecasts are only possible out to about 10-14 days \cite[e.g.,][]{stern2015trends}.  However, dynamical processes in the ocean operate on much longer time scales, and many atmospheric processes depend crucially on the ocean as a forcing mechanism.  This coupling between the slowly varying ocean and the faster varying atmosphere (and associated processes), allows for the skillful prediction of some general properties of the atmospheric state many months to over a year in advance (i.e., long-lead forecasting).  Although statistical models are not as skillful as deterministic numerical weather prediction models for short-to-medium weather forecasting, they have consistently performed as well or better than deterministic models for long-lead forecasting \cite[e.g.,][]{barnston1999predictive, jan2005did}.  In some cases, fairly standard linear regression or multivariate canonical correlation analysis methods can be used to generate effective long-lead forecasts \cite[e.g.,][]{penland1993prediction,knaff1997nino}.  However, given the inherent nonlinearity of these systems, it has consistently been shown that well crafted nonlinear statistical methods often perform better than linear methods, at least for some spatial regions and time spans \cite[e.g.,][]{drosdowsky1994analog,tang2000skill,berliner2000long,timmermann2001empirical,kondrashov2005hierarchy,wikle2010general,gladish2014physically}. It remains an active area of research to develop nonlinear statistical models for long-lead forecasting, and there is a need to develop methods that are computationally efficient, skillful, and can provide realistic uncertainty quantification in the presence of multiple time and spatial scales.

Statistical approaches for nonlinear dynamical spatio-temporal models (DSTMs) have focused on accommodating the quadratic nonlinearity that is present in many mechanistic models of such systems \citep[][]{kravtsov2005multilevel,wikle2010general,richardson2017sparsity}.  These models, at least when implemented in a way that fully accounts for uncertainty in data, process, and parameters, can be quite computationally challenging, mainly due to the very large number of parameters that must be estimated.  Solutions to this challenge require reducing the dimension of the state-space, regularizing the parameter space, the incorporation of additional information (prior knowledge), and novel computational approaches \citep[see the summary in][]{wikle2015modern}.  Parsimonious alternatives include analog methods \citep[e.g.,][]{mcdermott2016model,zhao2016analog}, and individual (agent) based models \citep[e.g.,][]{hooten2010statistical}. 

RNNs provide an alternative approach to model multiple scale nonlinear spatio-temporal processes. In essence, RNNs are a type of artificial neural network, originally developed in the 1980s \citep[][]{hopfield1982neural} that, unlike traditional feed-forward neural networks, include ``memory'' and allow cycles that can process sequences in their hidden layers.  That is, unlike feed-forward neural networks, RNNs explicitly account for the dynamical structure of the data. This has made them ideal for applications in natural language processing, speech recognition, and image captioning.  These methods have not been used extensively for spatio-temporal prediction, although there are notable exceptions \citep[][]{dixon2017deep,mcdermott2017bayesian}. Like the quadratic nonlinear spatio-temporal DSTMs in statistics, these models have a very large number of parameters (weights), and can be quite difficult to tune and train, and are computationally intensive.   One way to get the advantages of a RNN within a more parsimonious parameter estimation context is through the use of so-called ``reservoir computing'' methods - the most common of which is the ESN \citep[][]{jaeger2001echo}. In this case the hidden states and inputs evolve in a ``dynamical reservoir'' in which the parameters (weights) that describe their evolution are drawn at random with most (e.g., 90\% or upwards) assumed to be zero.  Then, the only parameters that are estimated are the output parameters (weights) that connect the hidden states to the output response.  For example, in the context of continuous spatio-temporal responses, this is just a regression estimation problem (usually with a regularization penalty; e.g., ridge regression).  See Section \ref{basicESN} for a model representation of the ESN.

Historically, the reservoir parameters in the ESN are just chosen once, with fairly large hidden state dimensions. Although this often leads to good predictions, it provides no opportunity for uncertainty quantification.  An alternative is to perform parametric bootstrap or ensemble estimation, in which multiple reservoir samples are drawn \citep[e.g.,][]{sheng2013prediction,mcdermott2017ensemble}.  This provides a measure of uncertainty quantification and allows one to choose smaller hidden state dimensions, essentially building an ensemble of weak learners analogous to many methods in machine learning \citep[e.g.,][]{friedman2001elements}.  There have also been Bayesian implementations of the ESN model \citep[e.g.,][]{li2012chaotic,chatzis2015sparse}, but none of these have been implemented with traditional Markov chain Monte Carlo (MCMC) estimation methods, as is the case here, where multiple levels of uncertainties can be accounted for.  In the context of spatio-temporal forecasting, \citep[][]{mcdermott2017bayesian,mcdermott2017ensemble} have shown that augmenting the traditional ESN with quadratic output terms (analogous to the quadratic nonlinear component in  statistical DSTMs) and input embeddings (e.g., including lags of the input variables as motivated by \citeauthor{takens1981detecting}' (\citeyear{takens1981detecting}) representation in dynamical systems) can improve forecast accuracy compared to traditional ESN models.  

``Deep'' models have recently shown great success in many neural network applications in machine learning \citep[e.g.,][]{krizhevsky2012imagenet}.  As mentioned above, deep (hierarchical) statistical models have also been shown to be effective in complex spatio-temporal models.  There are challenges in training such models given the very large number of parameters (weights), so it can be advantageous to consider deep models that are also relative parsimonious in their parameter space.  It is then natural to explore the potential for deep or hierarchical ESN models.  The purpose of adding additional layers in the ESN framework is to model (learn) additional temporal scales \citep[][]{jaeger2007discovering}. Deep ESN models provide a greater level of flexibility by allowing individual layers to potentially represent different time scales. The model presented here attempts to exploit the multiscale nature of deep ESN models for spatio-temporal forecasting.

Although deep ESN models have been considered in the engineering literature  \citep[][]{jaeger2007discovering,triefenbach2013acoustic,antonelo2017echo,ma2017deep}, none of these approaches accommodate uncertainty quantification.  Furthermore, these methods do not include a spatial component nor are they applied to spatio-temporal systems.  Here we develop a hierarchical ESN approach for spatio-temporal prediction that explicitly accounts for uncertainty quantification.  We first present an ensemble approach, extending the work of \cite{mcdermott2017ensemble} to the deep setting, and then consider a hierarchical Bayesian implementation of the deep ESN model that more rigorously accounts for observation model uncertainty.

Section 2 describes the motivating spatio-temporal long-lead forecasting problem, in this case using Pacific sea surface temperature (SST) anomalies to forecast soil moisture anomalies over the US ``corn belt'' 6 months in the future.  Section 3 describes the hierarchical ESN methodology, first from the ensemble perspective and then the Bayesian perspective.  Section 4 presents a simulation study using a non-Gaussian multiscale Lorenz-96 system and then presents the soil moisture forecasting example.  We conclude with a brief discussion in Section 5.

\section{Motivating Problem: Long-Lead Forecasting}

In the context of atmospheric and oceanic processes, {\it long-lead forecasting} refers to forecasting atmospheric, oceanic or related variables on monthly to yearly time scales. Although it is fundamentally impossible to generate skillful meteorological forecasts of atmospheric processes on time horizons of greater than about ten days to two weeks, the ocean operates on much longer time scales than the atmosphere and provides a significant amount of the forcing of atmospheric variability.  Thus, this linkage between the ocean and the atmosphere can lead to skillful long-lead ``forecasts'' of the atmospheric state, or other processes linked to the atmospheric state (e.g., soil moisture) on time scales of months to two years or so \cite[e.g.,][]{philander1990nino}.  In this case, one cannot typically generate skillful meteorological point forecasts, but can provide general distributional forecasts that are skillful relative to na\"{i}ve models such as climatology (long-term averages) or persistence (assuming current conditions persist into the future).  

Historically, successful long-lead forecasting applications are typically tied to the ocean El Ni\~no/Southern Oscillation (ENSO) phenomenon, which shows quasi-periodic variability between warmer than normal ocean states in the central and eastern tropical Pacific ocean (El Ni\~no) and colder than normal ocean states in the central tropical Pacific (La Ni\~na).  The ENSO phenomenon accounts for the largest amount of variability in the tropical Pacific ocean and leads to world wide impacts due to atmospheric ``teleconnections'' (i.e., the shifting of the warm pools in the tropical Pacific also shift the convective clusters of precipitation that drive upper atmospheric wave trains, that in turn influence the atmospheric circulation; e.g., locations of jet streams). These atmospheric circulation changes then affect temperature, precipitation, and many responses to those variables such as habitat conditions for ecological processes, soil moisture, severe weather, etc., as described in \citet[][]{philander1990nino}.  It has been demonstrated for over two decades that skillful long-lead forecasts of sea surface temperature (SST) in the tropical Pacific (and, hence, ENSO) is possible with both deterministic and statistical models. In fact, this is one of the aforementioned situations in the atmospheric and ocean sciences where statistical forecasting is as good as, or better than, deterministic models \cite[e.g.,][]{barnston1999predictive, jan2005did}.

There are essentially two general approaches to the long-lead forecasting of a response to SST forcing.  One approach is to generate a long-lead forecast of SST (e.g., a 6 month forecast) and then use contemporaneous relationships between the ocean state and the response of interest (e.g., say midwest soil moisture) to generate a long-lead forecast of that response.  This typically requires a dynamical forecast of SST and then some regression or classification model from SST to the outcome of interest.  The alternative is to model the relationship between SST and the future response at a chosen lead time (e.g., forecasting midwest soil moisture in May given SST in November).  This might be called a {\it spatio-temporal regression}, where one is predicting a spatial field in the future given a spatial field at the current time.   In either case, linear models have been shown to perform reasonably well in these situations \cite[e.g.,][]{van2003performance}, but it is typically the case that models that include nonlinear interactions can perform more skillfully, and can produce more realistic long-lead forecast distributions \cite[e.g.,][]{sheffield2004simulated,fischer2007soil,mcdermott2016model}.  

\subsection{Long Lead Forecasting of Soil Moisture in the Midwest U.S. Corn Belt}

Soil moisture is fundamentally important to many processes (e.g., agricultural production, hydrological runoff).  In particular, the amount of soil moisture available to crops such as wheat and corn at certain critical phases of their growth cycle can make a significant impact on yield \citep[][]{carleton2008synoptic}.  Thus, having a long-lead understanding of the plausible distribution of soil moisture over an expansive area of agricultural production can assist producers by suggesting optimal management approaches (e.g., timing of planting, nutrient supplementation, and irrigation).   Given the aforementioned links between tropical Pacific SST and North American weather patterns, it is not surprising that skillful long-lead forecasts of soil moisture in major production areas of the U.S.  are possible \cite[e.g.,][]{van2003performance}.  Indeed, the U.S. National Oceanic and Atmospheric Administration (NOAA) and National Center for Environmental Prediction (NCEP) routinely provide soil moisture outlooks (``forecasts") that are based on a combination of deterministic and statistical (constructed analog) models (e.g., see \verb+ http://www.cpc.ncep.noaa.gov/soilmst+ \\ \verb+/index_jh.html+), although for shorter lead times than of interest here.  Recently,  \citet[][]{mcdermott2016model} showed that a Bayesian analog forecasting model could provide skillful high-spatial-resolution forecasts of soil moisture anomalies over the state of Iowa in the U.S. at lead times up to 6 months. 

The application here will consider the problem of forecasting soil moisture over the midwest U.S. corn belt in May given data from the previous November (i.e., a 6 month lead time).  We consider May soil moisture because it is an important time period for planting corn in this region \citep[i.e.,][]{blackmer1989correlations}.   This application corresponds to the long-lead spatio-temporal field regression approach to nowcasting described above -- that is, we regress a tropical Pacific SST field in November onto the soil moisture field from the following May.  The details associated with the data and model used for this example are given in Section \ref{sec:SoilMoistureEx}.

\section{Methodology}

Suppose we are interested in forecasting the spatio-temporal process $\bZ_t \equiv (Z_t(\bs_1),\ldots,Z_t(\bs_{n_z}))'$ for time periods $\{t=1,\dots,T\}$ at a discrete set of spatial locations $\{\bs_i \in D_z \subset \R^2: i=1,\dots,n_z\}$. Using a chosen linear dimension reduction method, $\bZ_t$ can be decomposed such that $\bZ_t \approx {\mbv \Phi} {\mbv \alpha}_t$, where ${\mbv \Phi}$ is a $n_z \times n_b$ matrix of $n_b$ spatial basis functions and ${\mbv \alpha}_t$ is a $n_b$-dimensional vector of basis coefficients, indexed by $b=1,\dots,n_b$ \citep[e.g.,][Chap. 7]{{CandW2011}}. Next, assume we have inputs corresponding to a spatio-temporal data set. Specifically, let $\bx_t \equiv (x_t(\br_1),\ldots,x_t(\br_{n_x}))'$ be a vector of $n_x$ input variables that correspond to a discrete set of spatial locations $\{\br_d \in D_z \subset \R^2: d=1,\dots,n_x\}$ at time $t$.  (More generally the input vector may consist of $n_x$ time-varying input covariates that are not indexed by space; e.g., $\{\br_d \equiv r_d \subset R_d: d=1,\dots,n_x\}$).

Below we provide a brief overview of the ensemble ESN model considered in \cite{mcdermott2017ensemble}, followed by the multi-level (deep) extension that we call a deep ensemble echo state network (D-EESN) model.  We then describe a Bayesian version of the D-EESN model that we label (BD-ESSN).  

\subsection{Basic ESN Background}\label{sec:basic_ESN}
\label{basicESN}

The quadratic ESN model outlined in \cite{mcdermott2017ensemble} is given by:
\begin{eqnarray}
 \mbox{\bf Data stage:} & \; &   \;\;   \;\;  \bZ_t \approx {\mbv \Phi} {\mbv \alpha}_t  \label{eq:QESNdata} \\ 
\mbox{\bf Output stage: }  & \; & {\bfalpha}_t = {\mbf V}_1 {\mbf h}_t + {\mbf V}_2 {\mbf h}^2_t + {\mbv \eta}_t, \;\; {\mbv \eta}_t \; \sim \; \text{Gau}({\mbf 0},\sigma^2_\eta {\mbf I} ) \label{eq:QESNoutput} \\
\mbox{\bf Hidden stage: } & \; &  \bh_t = g_h\left(\frac{\nu}{|\lambda_w|}{\mbf W} {\mbf h}_{t-1} + {\mbf U}\tilde{\mbf x}_t\right), \label{eq:QESNhidden}   
\end{eqnarray}
where $\bfPhi$ is an $n_z \times n_b$ matrix of spatial basis functions, $\bfalpha_t$ is a $n_b$-vector that contains the associated basis expansion coefficients (where $n_b << n_z$), $\bh_t$ is an $n_h$-dimensional vector of ``hidden units,'' $g_h(\cdot)$ is a nonlinear activation function (e.g., a sigmoidal function such as a hyperbolic tangent function), $\lambda_w$ is the ``spectral radius'' (largest eigenvalue) of $\bW$, $\nu$ is a scaling parameter, and $\bW$, $\bU$, $\bV_1$, $\bV_2$ are weight (parameter) matrices of dimension $n_h \times n_h$, $n_h \times n_{\tilde{x}}$, $n_b \times n_h$, and $n_b \times n_h$, respectively (defined below). The square parameter matrix $\bW$ can be thought of analogously to a transition matrix in a vector autoregressive (VAR) model in that it can capture transition dynamic interactions between various inputs. The scaling parameter $\nu$ helps control the amount of memory in the system and is restricted such that, $0\le \nu \le 1$ for stability purposes \citep[][]{jaeger2007discovering}. We let $\tilde{\bx}$ be an $n_{\tilde{x}}$-vector of $m$ embeddings (lagged input values) given by:
\begin{equation}
\tilde{\mbf x}_t = [\bx'_t,\bx'_{t-\tau}, \bx'_{t- 2 \tau},\ldots,\bx'_{t - m \tau}]' \label{eq:QESNembed}.
\end{equation}

Importantly, only $\sigma^2_\eta$ and the weight matrices $\bV_1$ and $\bV_2$ in the output stage (\ref{eq:QESNoutput}) are estimated (usually with a ridge penalty). These weight matrices can be thought of similarly to regression parameters that weight the hidden units appropriately.   In contrast, the  ``reservoir weights'' in (\ref{eq:QESNhidden}) are drawn from the following distributions:
\begin{eqnarray}
& \; & {\mbf W} = [w_{k,q}]_{k,q}: w_{k,q} = \gamma^w_{k,q} \; \text{Unif}(-a_w,a_w) + (1 - \gamma^w_{k,q}) \; \delta_0, \label{eq:QESNw} \\
& \; & {\mbf U} = [u_{k,r}]_{k,r}: u_{k,r} = \gamma^u_{k,r} \; \text{Unif}(-a_u,a_u) + (1 - \gamma^u_{k,r}) \; \delta_0, \label{eq:QESNu}  \\
& \; & \gamma_{k,q}^w \; \sim \; Bern(\pi_w), \hspace{.25cm} \gamma_{k,r}^u \; \sim \; Bern(\pi_u), 
\end{eqnarray}
where $\gamma_{k,q}^w$ and $ \gamma_{k,r}^u$ denote indicator variables, $\delta_0$ denotes a Dirac function, and $\pi_w$ (and $\pi_u$) can be thought of as the probability of including a particular weight (parameter) in the model. As is common in the machine learning literature, both  $a_w$ and $a_u$ are also set to small values to help prevent overfitting. Similarly, the parameters $\pi_w$ and $\pi_u$ are set to small values to create a sparse network.  Both the sparseness and randomness act as a regularization mechanism, which prevents the ESN model from overfitting to in-sample data.  In summary, note that the hidden (reservoir) stage in (\ref{eq:QESNhidden}) corresponds to a {\it nonlinear stochastic transformation of the input vectors} that are then regressed, with regularization, onto the output vectors in the output stage (i.e., (\ref{eq:QESNoutput}) above), which are then transformed back to the data scale in (\ref{eq:QESNdata}).

Traditional ESN models \citep[e.g.,][]{jaeger2007discovering,lukovsevivcius2009reservoir} do not typically include the quadratic ($\bV_2$) term in the output stage nor do they include the embeddings in the input vector, but \cite{mcdermott2017ensemble} found that those are often helpful when forecasting spatio-temporal processes.  In addition, traditional ESN applications typically include a  ``leaking rate'' parameter that corresponds to a convex combination of the previous hidden state $\bh_{t-1}$ and the current reservoir value, $\bh_t$ \citep[e.g.,][]{lukovsevivcius2012practical}.  We have not found this to be as useful for long-lead spatio-temporal forecasting as it has been in other applications and so omit it in our presentation for notational simplicity. 

To facilitate uncertainty quantification, \cite{mcdermott2017ensemble} were motivated by parametric bootstrap prediction methods \citep[e.g.,][]{genest2008validity,sheng2013prediction} to consider an ensemble of predictions from the quadratic ESN model.  In particular, Algorithm 1 of \cite{mcdermott2017ensemble} provides a simple procedure that generates $n_{res}$ forecast realizations $\{\widehat{\bZ}^{(j)}_t = \bfPhi \widehat{\bfalpha}^{(j)}_t$: $j=1,\ldots,n_{res}\}$ by implementing the reservoir model in (\ref{eq:QESNhidden}) to obtain $\bfalpha^{(j)}_t$ in (\ref{eq:QESNoutput}) through the use of $n_{res}$  simulated (sample) weight matrices from (\ref{eq:QESNw}) and (\ref{eq:QESNu}).  These samples can then be used in a Monte Carlo sense to obtain features of the predictive distribution (e.g., means, variances, etc.). This quadratic-ensemble ESN (Q-EESN) model was shown to be quite effective in capturing the uncertainty associated with long-lead forecasting of tropical Pacific SST.

\subsection{Deep Ensemble ESN (D-EESN)}\label{sec:D-EESN}

Here we develop a deep extension of the bootstrap-based Q-EESN model of \cite{mcdermott2017ensemble} described in Section \ref{sec:basic_ESN}.  That model has multiple levels, but is not a ``deep'' model in the sense that it has no mechanism to link hidden layers, which might be important for processes that occur on multiple time scales.   To accommodate such structure, we extend some of the deep ESN model components developed in \cite{ma2017deep} and \cite{antonelo2017echo} to a spatio-temporal ensemble framework in the following D-EESN model.  In particular, the D-EESN model with $\ell=1,\dots,L$ hidden layers is defined as follows for time period $t$ and $L\ge2$:

\vspace{-10mm}

\begin{eqnarray}
 & \; & \text{ \bf Data Stage:}   \;\;   \;\;  \bZ_t \approx {\mbv \Phi} {\mbv \alpha}_t  \label{eq:D-EESNdata} \\ 
 & \; & \text{ \bf Output Stage:}  \;\;   \;\;   {\mbv \alpha}_t =\bV_1 \bh_{t,1}+ \sum\limits_{\ell=2}^L \bV_\ell g_h(\widetilde{\bh}_{t,\ell})   + {\mbv \eta_t}, \hspace{.1cm}   {\mbv \eta_t}\sim \text{Gau}(\mbv{0},\sigma^2_\eta \bI ),  \label{eq:D-EESNoutput} \\
 & \; &   \hspace{3.15cm}  \text{s.t.} \  \bv_{\ell,b}'  \bv_{\ell,b} \le c_v, \nonumber \\
  & \; & \text{ \bf Hidden Stage $\ell$:}   \;\;   \;\;   \bh_{t,\ell}=f \Big(  \frac{\nu_\ell}{ \lambda_{W_\ell}} \bW_\ell \bh_{t-1,\ell} + \bU_\ell \widetilde{\bh}_{t,\ell+1}  \Big),  \;\;   \text{for}  \;\ \ell < L \label{hiddenStageOne} ,\\
    & \; & \text{ \bf Reduction Stage $\ell+1$:}    \;\;   \;\;   \widetilde{\bh}_{t,\ell+1} \equiv \mathcal{Q}(\bh_{t,\ell+1}), \;\;   \text{for}  \;\ \ell < L, \label{eq:DEESNredStage} \\
    & \; & \text{ \bf Input Stage:}  \;\;   \;\;  \;\;   \;\;   \bh_{t,L}=f \Big(  \frac{\nu_L}{ \lambda_{W_L}} \bW_L \bh_{t-1,L} + \bU_L \widetilde{\bx}_t  \Big) \label{eq:D-EESNinput},
\end{eqnarray}
where $\bh_{t,\ell}$ is a $n_{h,\ell}$-dimensional vector for the $\ell^{th}$ hidden layer and each parameter matrix $\bV_\ell$ is defined as:
 \begin{align}
    \bV_\ell & \equiv\begin{bmatrix}
           \bv_{\ell,1}' \\
           \vdots \\
         \bv_{\ell,n_b}'
         \end{bmatrix}.
  \end{align}
Let $\bh_{1:T,\ell+1} \equiv [\bh_{1,\ell+1},\ldots,\bh_{T,\ell+1}]'$ be a $T \times n_{h,\ell+1}$ matrix. The function $\mathcal{Q}(\cdot)$ in (\ref{eq:DEESNredStage}) denotes a dimension reduction mapping function of this matrix (see below for specific examples) such that $\mathcal{Q}: \bh_{1:T,\ell +1} \rightarrow \widetilde{\bh}_{1:T,\ell +1} $ for $\widetilde{\bh}_{1:T,\ell +1}\equiv\mathcal{Q}(\bh_{1:T,\ell +1} )$, where $\widetilde{\bh}_{1:T,\ell +1}$ is a $T \times n_{\tilde{h},\ell +1}$ matrix such that $n_{h,\ell +1}\ge n_{\tilde{h},\ell +1}$. More concisely, this transformation is on the matrix $\bh_{1:T,\ell +1}$, resulting in the transformed matrix $\widetilde{\bh}_{1:T,\ell +1}$, and then for each time period the appropriate $\widetilde{\bh}_{t,\ell +1}$ is extracted from the transformed matrix to construct (\ref{hiddenStageOne}). Through the inclusion of the $\widetilde{\bh}_{t,\ell}$ terms in (\ref{eq:D-EESNoutput}), the model can potentially weight layers lower down in the hierarchy that may represent different time scales or features \citep[i.e.,][]{ma2017deep}. Note, the activation function $g_h(\cdot)$ is included in (\ref{eq:D-EESNoutput}) to place the dimension reduction variables on a similar scale as the $\bh_{t,1}$ variables (i.e., the hidden variables that have not gone through the dimension reduction transformation).

For a given hidden unit $\ell$, $\lambda_{w_\ell}$ denotes the largest eigenvalue of the square matrix $\bW_\ell$, where as in the basic ESN model above, $\bW_\ell= [w^{(\ell)}_{k_\ell,q_\ell}]_{k_\ell,q_\ell}$ is drawn from the following distribution:
\begin{equation}
 w_{k_\ell,q_\ell}^{(\ell)} = \gamma^{w_\ell}_{k_\ell,q_\ell} \; \text{Unif}(-a_{w_\ell},a_{w_\ell}) + (1 - \gamma^{w_\ell}_{k_\ell,q_\ell}) \; \delta_0, \hspace{.5cm} \gamma_{k_\ell,q_\ell}^{w_\ell} \; \sim \; Bern(\pi_{w_\ell}), \label{w_DEESN}\\
\end{equation}
and each element of the parameter matrix $\bU_\ell=[u^{(\ell)}_{k_\ell,r_\ell}]_{k_\ell,r_\ell}$ is drawn from:
\begin{equation}
 u_{k_\ell,r_\ell}^{(\ell)} = \gamma^{u_\ell}_{k_\ell,r_\ell} \; \text{Unif}(-a_{u_\ell},a_{u_\ell}) + (1 - \gamma^{u_\ell}_{k_\ell,r_\ell}) \; \delta_0, \hspace{.5cm} \gamma_{k_\ell,r_\ell}^{u_\ell} \; \sim \; Bern(\pi_{u_\ell}). \label{u_DEESN} \\
\end{equation}
The embedded input vector, $\widetilde{\bx}_t $ is defined as in (\ref{eq:QESNembed}) above.  Estimation of $\bV_1,\dots,\bV_L$ is carried out through the use of ridge regression using the ridge hyper-parameter $r_v$, where there is a one-to-one relationship between $r_v$ and the constant $c_v$ in (\ref{eq:D-EESNoutput}).  Note that we do not include a quadratic output term in this model as we did in (\ref{eq:QESNoutput}) for simplicity, but this could easily be added if the application warrants (note -  both applications presented here did not benefit from the addition of quadratic output terms in the deep model setting).

The  bootstrap ensemble prediction for the D-EESN model is presented in Algorithm \ref{algorithm1} below. In particular, Algorithm \ref{algorithm1} starts by drawing $\bW_\ell$ and $\bU_\ell$ for every layer in the D-EESN model. Next, these parameters are plugged-in sequentially, starting with the input layer defined in (\ref{eq:D-EESNinput}) and ending with the hidden layer that comes directly before the output stage (i.e., (\ref{hiddenStageOne}) for $\ell=1$). Finally, with all of the hidden states calculated, ridge regression is used to estimate the regression parameters in (\ref{eq:D-EESNoutput}).


\vspace{.5cm}

\begin{spacing}{1.25}
\begin{algorithm}[H]
  \caption{D-EESN algorithm}
        {\bf Data :} $\{\mbv \alpha_t, \tilde{\mbf x}_t : t=1,\dots, T \}$ \\
      {\bf Input :} hyper-parameters: $\{ \nu_1,\dots,\nu_L,n_{\tilde{h},2},\dots, n_{\tilde{h},L}, n_{h,1},r_v\}$, chosen by cross-validation. \\
    \For {$j=1, \dots, n_{res}$} {
    Simulate  $ \bW_\ell$ and $\bU_\ell $ from (\ref{w_DEESN}) and (\ref{u_DEESN}) for $\ell=1,\dots,L$ \\
    Calculate $\{\bh_{t,L}: t=1,\dots, T\}$ with (\ref{eq:D-EESNinput}) using $\bW_L $ and $\bU_L $ \\
        \For {$\ell=L-1$ to $1$} {
         Calculate  $\{\widetilde{\bh}_{t,\ell+1}: t=1,\dots, T\}$  using (\ref{eq:DEESNredStage}) \\
          Calculate  $\{\bh_{t,\ell}: t=1,\dots, T\}$  with (\ref{hiddenStageOne}) using $\bW_\ell $ and $\bU_\ell $ \\
	}
	Use ridge regression to calculate $\bV_1^{(j)},\dots, \bV_L^{(j)}$ \\
	Calculate out-of-sample forecasts $\{\widehat{\bZ}^{(j)}_t: t=T+1,\dots, T+n_f\}$ \\
    }
    {\bf Output :} Ensemble of forecasts $\{\widehat{\bZ}^{(j)}_t: t=T+1,\dots, T+n_f\ ; j=1,\dots,n_{res} \}$  \\~\\
     \label{algorithm1}
\end{algorithm}
\end{spacing}

\vspace{.85cm}

The reservoir hyper-parameters, $\{ \nu_1,\dots,\nu_L,n_{\tilde{h},2},\dots, n_{\tilde{h},L}, n_{h,1},r_v\}$ are chosen by cross-validation driven by a genetic algorithm (see Section \ref{modelSetup} for further details).  As in \cite{mcdermott2017ensemble}, we found that fixing the parameters 
$\{\ \pi_{w_1},\dots,\pi_{w_L}, \pi_{u_1} \dots, \pi_{u_L}, a_{w_1}, \\ \dots, a_{w_L}, a_{u_1}, \dots, a_{u_L}  \}$ as well as the number of hidden units for all of the layers except the first (i.e., $ \{ n_{h,2}, \dots, n_{h,L} \}$) was a reasonable assumption here since the dimension of all the hidden units besides the top layer (i.e., $n_{h,1}$) are eventually reduced using the dimension reduction transformation. 

Note that this model consists of a series of {\it linked non-linear stochastic transformations of the input vector} that are available for prediction.  In addition, the data reduction steps act similarly to the ``pooling'' step in a convolutional neural network \citep[][]{krizhevsky2012imagenet}.  That is, it simultaneously reduces the dimension of the hidden layers and provides a summary of the important features. Similar to how ridge regression acts to reduce redundancy in the classic ESN model, dimension reduction serves this purpose in the deep framework. Due to the limited amount of research on deep ESNs, the question of which dimension reduction method to use for $\mathcal{Q}(\cdot)$ in (\ref{eq:DEESNredStage}) is still an open research question. Here, principal component analysis (PCA) and Laplacian eigenmaps \citep[][]{laplacianEigen} were explored as possible choices for $\mathcal{Q}(\cdot)$. Although these methods were selected to represent both linear and nonlinear dimension reduction techniques, there are certainly other choices that could be explored in future applications.

\subsection{Bayesian Deep Ensemble ESN (BD-EESN) } \label{sec:BD-EESN}

The D-EESN model given in Section \ref{sec:D-EESN} is very efficient to implement, but at a potential cost of not fully accounting for all sources of uncertainty in estimation and prediction.  In particular, the data stage given in (\ref{eq:D-EESNdata}) does not account for truncation error in the basis expansion, nor the error associated with the estimates of the regression or residual variance parameters in (\ref{eq:D-EESNoutput}).  This can easily be remedied via Bayesian estimation at these stages. To our knowledge, this is the first ESN model for spatio-temporal data to be implemented within a traditional hierarchical Bayesian framework.

We develop the Bayesian deep EESN (BD-EESN) model in a general form here so that it can be applied to both the traditional EESN and the D-EESN described above. In particular, let:
\begin{eqnarray}
 & \; & \text{ \bf Data stage:}  \;\;   \;\;   {\mbf Z}_t | \bfalpha_t  \; \sim \; \text{D} ( {\mbv \mu}( {\mbv \alpha}_t), {\mbv \Theta} ),  \label{eq:BD-EESNdata}  \\
 & \; &  \text{\bf Output stage:} \;\;   \;\;   {\mbv \alpha}_t =\frac{1}{n_{res}} \sum\limits_{j=1}^{n_{res}} \big[  {\mbv \beta}_1^{(j)}\bh_{t,1}^{(j)} +  \sum\limits_{\ell=2}^L {\mbv \beta}_\ell^{(j)}  g_h(\widetilde{\bh}_{t,\ell}^{(j)}) \big] +{\mbv \eta}_t,  \label{eq:BD-EESNoutput}
 \end{eqnarray}
 where ${\mbv  \eta}_t  \sim \text{Gau}({\mbf 0},\sigma^2_ \eta \bI)$ and D denotes an unspecified distribution; for both of the applications discussed here $ {\mbv \mu}( {\mbv \alpha}_t)\equiv{\mbv \Phi} {\mbv \alpha}_t$ and ${\mbv \Theta} \equiv{\mbv \Sigma}_z$ (this specification is application dependent and could be altered to accommodate other distributions). Here, ${\mbv \Sigma}_z$ is defined as a (known) $n_z\times n_z$ spatial covariance matrix (e.g., accommodating the basis truncation and/or measurement error), see Section \ref{results} for application-specific details. The regression matrices in (\ref{eq:BD-EESNoutput}) are defined as follows for $\ell=1,\dots, L$ and $j=1,\dots,n_{res}$:
  \begin{align}
    {\mbv \beta}_\ell^{(j)}& \equiv\begin{bmatrix}
         {\mbv \beta}_{\ell,1}^{(j)'}\\
           \vdots \\
   {\mbv \beta}_{\ell,n_b}^{(j)'}
         \end{bmatrix}.
          \label{betaDef}
  \end{align}
In this notation, $\bh_{t,1}^{(j)}$ is the $j^{th}$ sampled $n_{h,1}$-dimensional vector from a $n_{h,1} \times T$ dynamical reservoir generated as in the D-EESN model in Section \ref{sec:D-EESN}; that is, we generate $j=1,\ldots,n_{res}$ reservoir samples ``off-line'' using (\ref{hiddenStageOne}) from the D-EESN model (for $\ell=1$). Similarly,  $\widetilde{\bh}^{(j)}_{t,2}, \dots, \widetilde{\bh}^{(j)}_{t,L}$ are also sampled {\it a priori}  and come from the dimension reduction stage(s) of the D-EESN model (i.e., (\ref{eq:D-EESNinput}) above). That is, both $\bh_{t,1}^{(j)} $ and $\{\widetilde{\bh}_{t,\ell}^{(j)}: \ell=2,\dots, L\}$ are treated as fixed covariates for the BD-EESN model. So, the output model takes an ensemble approach in terms of generating a suite of nonlinear stochastic transformed input variables for a Bayesian regression, where each $\{ {\mbv \beta}^{(j)}_\ell: \ell=1,\dots, L \}$ are matrices of regression parameters as defined in (\ref{betaDef}). Note the obvious similarity between the process model in (\ref{eq:D-EESNoutput}) and the one for the D-EESN model in (\ref{eq:BD-EESNoutput}). If all of the ${\mbv \beta}_\ell^{(j)} $ terms for $\ell \ge 2$ are set to zero, the process model in  (\ref{eq:BD-EESNoutput}) can be used with the traditional EESN model (or the Q-EESN model by simply adding quadratic terms).


This model is clearly over-parameterized, so the regression parameters in the BD-EESN model are given stochastic search variable selection (SSVS) priors \citep[][]{george1997approaches}. While one of the many variable selection priors from the Bayesian variable selection literature can be used here, we choose to use SSVS priors for their ability to produce efficient shrinkage. In particular, SSVS priors shrink a (large) percentage of parameters to zero (in a spike-and-slab implementation) or infinitesimally close to zero, while leaving the remaining variables unconstrained. Thus, the regression parameters in the BD-EESN model are given the following hierarchical prior distribution for $\ell =1,\ldots,L$ and $ j=1,\dots, n_{res}$: 
 \begin{eqnarray}
 &\; &   \;\;   \;\;  \beta^{(j)}_{\ell,b,k_\ell} \mid \gamma_{\ell}^{\beta_\ell} \sim \gamma_{\ell}^{\beta_\ell} \;  \text{Gau}(0,\sigma^2_{\beta_\ell,0}) +  (1- \gamma_{\ell}^{\beta_\ell} ) \; \text{Gau}(0,\sigma^2_{\beta_\ell,1}),  \label{eq:BD-EESNbeta} \\
 &\;& \;\; \;\; \gamma_{\ell}^{\beta_\ell}\sim \text{Bernoulli}(\pi_{\beta_\ell}),   \nonumber 
 \end{eqnarray}
where $k_\ell$ indexes the hidden units for a particular layer, $\sigma^2_{\beta_\ell,0} >> \sigma^2_{\beta_\ell,1}$, and $\pi_{\beta_\ell}$ can be thought of as the prior probability of including a particular variable in the model. Finishing the prior specifications for the model, the variance parameter $\sigma^2_\eta $ is given an inverse-gamma prior such that $\sigma^2_\eta \sim \text{IG}(\alpha_\eta,\beta_\eta)$. The hyper-parameters $\sigma^2_{\beta_\ell,0}$, $\sigma^2_{\beta_\ell,1}$, $\pi_{\beta_\ell}$, $\alpha_\eta$, and $\beta_\eta$ are problem-specific (see the examples in Section \ref{modelSetup} below). 

Since the parameters here are given conjugate priors, the full-conditional distributions that make up the Gibbs sampler MCMC algorithm needed to implement this model are straightforward to sample from. A summary of the entire estimation procedure for the BD-EESN can be found in Algorithm \ref{algorithm2}. Forecasts for the BD-EESN are made in a similar manner as the D-EESN (as shown by the output step of Algorithm \ref{algorithm1}). That is, training is carried out using data up to time period $T$, while out-of-sample forecasts are continually made $\tau$ periods into the future using the corresponding lagged input variable to generate the appropriate hidden state variables. Finally, at each iteration of the MCMC algorithm, the sampled regression parameters are used with these hidden states to produce out-of-sample forecasts.

\vspace{.5cm}

\begin{algorithm}[H]
\singlespace
\begin{enumerate}
\item Use Algorithm \ref{algorithm1} from the D-EESN model with a genetic algorithm-based cross-validation to pick the reservoir hyper-parameters that make up the D-EESN model (i.e., the input for Algorithm \ref{algorithm1}). 
\item Using the hyper-parameters selected in Step 1, generate $n_{res}$ reservoirs from the hidden layers of the  D-EESN model to be used in the Bayesian model as covariates.
\item Use Gibbs sampling to estimate ${\mbv \alpha}_{1:T},  {\mbv \beta}_1^{(1:n_{res})}, \dots,{\mbv \beta}_L^{(1:n_{res})},$ and $ \sigma^2_\eta$, while treating the $n_{res}$ reservoirs generated in Step 2 as (fixed) covariates.
\end{enumerate}
\caption{Outline of estimation procedure for the BD-EESN.}
\label{algorithm2}
\end{algorithm}

\vspace{.85cm}

\section{Simulation and Motivating Examples}
\label{results}
Here we describe the model setup used to implement the D-EESN and BD-EESN models on a complex simulated process and for the soil moisture long-lead forecasting problem that motivated these models.

\subsection{Model Setup}
\label{modelSetup}
The previously mentioned cross-validation for the D-EESN is carried out using a genetic algorithm (GA) \citep[e.g.,][]{sivanandam2007introduction} contained in the \verb+GA+ package (\verb+https://cran.r-project.org/web/packages/GA+) from the R statistical computing program ( \verb+http://cran.r-project.us.org+). Unlike the basic ESN model, the number of hyper-parameters for the deep EESN model can increase quickly as the number of layers increases (e.g., with five-layers and a relatively coarse search grid the number of total parameters in the search space can easily approach $10^6$), thus rendering grid search approaches computationally burdensome at best \citep[i.e.,][]{ma2017deep}. Through the use of a GA, the hyper-parameters that make up a deep ESN can be selected at a fraction of the computational cost. The GA was implemented with 40 generations and a population size of 20 for all of the applications presented below. This is the first implementation, to our knowledge, of either a traditional or deep ESN within an ensemble framework using a GA. The bounds of the parameter search space for each of the hyper-parameters in the D-EESN model can be found in Table \ref{tab:Table_1}.

All ensemble models are comprised of 100 ensembles, we did not find any of the applications to be overly sensitive to this choice. Using this number of ensembles represents a compromise between computational efficiency and achieving consistent (reproducible) results (e.g., we found that using less than approximately 30 ensembles produced unstable results, while values greater than 100 did not change the results significantly). Note, previous bootstrap ensemble ESN papers have generally used far fewer ensembles \citep[e.g.,][]{sheng2013prediction}. For context, on a 2.3 GHz laptop the D-EESN algorithm defined in Algorithm \ref{algorithm1} takes 4.3 and 17.5 seconds for a two-layer and seven-layer model, respectively, using 100 ensembles with the Lorenz-96 application described in Section \ref{deepLorenz96} below. 

Regarding the previous discussed dimension reduction function for the reduction stage of the D-EESN model (i.e., (\ref{eq:DEESNredStage}) above), PCA basis functions were selected for both applications and models using cross-validation. Although, we should note that the model was not very sensitive to this choice among the basis functions considered.  Finally, for simplicity we use the same dimension (i.e., $n_{\tilde{h},\ell}$) for each dimension reduction layer in the D-EESN model, a similar assumption was made in \cite{ma2017deep}. Similar to \cite{mcdermott2017ensemble} all of the hyper-parameters in the set $\{\ \pi_{w_1},\dots,\pi_{w_L}, \pi_{u_1}, \dots, \pi_{u_L}, a_{w_1}, \\ \dots, a_{w_L}, a_{u_1}, \dots, a_{u_L}  \}$ are fixed at $0.10$. Finally, $n_{h,\ell}$ for $\ell \ge 2$ was set to 84 for all of the deep models. The model was not sensitive to this value with values ranging between 60 and 100 producing nearly identical results for the metrics evaluated below. As previously discussed, the bootstrap framework presented here aims to create an ensemble of weak learners, thus the selection of a moderate value for $n_{h,\ell}$.

\begin{table}[H]
\centering
\begin{tabular}{|c|ccccc|} 
\hline
 Hyper-Parameter: & $m$ & $\nu_\ell$ & $n_{\tilde{h},\ell}$  & $n_{h,1}$  &  $r_v$  \\
\hline
Search Space: & $\{0,1,\dots, 5 \}$ & $[0,1]$ & $\{6,7,\dots, 20 \}$ & $\{25,26,\dots, 75 \}$ & $[.0001,.01]$   \\ 
\hline
\end{tabular}
\caption{Search spaces for all of the hyper-parameters in the D-EESN model. These search spaces were implemented for the D-EESN using a genetic algorithm (GA) algorithm with cross-validation. Brackets $\{ \ \}$ denote discrete spaces and $[ \ ]$ denote continuous spaces.}
\label{tab:Table_1}
\end{table}

Estimation of all the Bayesian models considered here is carried out using a Gibbs sampler MCMC algorithm with 5,000 iterations where the first 1,000 iterations are treated as burn-in. The trace plots showed no evidence of  non-convergence (more details regarding convergence are given below). The specific hyper-parameters used for the Bayesian implementation can be found in Table \ref{tab:Table_2}. Note, more restrictive priors (in terms of regularization) are employed for the models with more regression parameters (i.e., models with more hidden layers). This improves the mixing of the MCMC algorithm along with preventing the model from overfitting. 

 \begin{table}[H]
\centering
 \hskip-.5cm
 \small
\begin{tabular}{ |c|l|l| }
\hline
Application: &   \multicolumn{1}{|c|}{Priors}   \\  [-.01cm]
 \hline
 BD-EESN models with $L=2$ & $\pi_{\beta_\ell}=.25, \sigma^2_{\beta_\ell,0}=5, \sigma^2_{\beta_\ell,1}=.001, \alpha_\eta=1$, and $\beta_\eta=1$  \  for $\ell=1, 2$ \\    [-.01cm]
 BD-EESN models with $L > 2$ & $\pi_{\beta_\ell}=.10, \sigma^2_{\beta_\ell,0}=4, \sigma^2_{\beta_\ell,1}=.001, \alpha_\eta=1$, and $\beta_\eta=1$  \  for $\ell=1,\dots, L$ \\   
  \hline
\end{tabular}
\caption{Specific hyper-parameters used for the various BD-EESN implementations. The hyper-parameters were selected so the in-sample MSE of the BD-EESN roughly matched the in-sample MSE for the corresponding D-EESN model. None of the applications were overly sensitive to the specified priors, with moderate variations from the values given below producing similar results.}
\label{tab:Table_2}
\end{table}

Both the D-EESN and BD-EESN are compared against the previously described single-layered Q-EESN model in order to investigate the added utility of using a deep framework. In addition, na\"{i}ve or simple forecasting methods are often employed for difficult long-lead forecasting problems such as the soil moisture application, to act as a baseline for comparison. We consider both a climatological and linear dynamical spatial-temporal model (DSTM) here as baseline models for the soil moisture application (for consistency, we also compare to the linear model in the deep Lorenz-96 simulation study described below). The linear DSTM model is defined as follows:
\begin{eqnarray}
  &\; &   \;\;   \;\;  {\mbf Z}_t \; \sim \; \text{Gau} ( {\mbv \Phi} {\mbv \alpha}_t, {\mbv \Sigma}^{(L)} ), \\
 &\; &   \;\;   \;\;    {\mbv \alpha}_t =\bM  {\mbv \alpha}_{t-1} + {\mbv \eta}^{(L)}_t,
 \end{eqnarray}
where $\bM$ is a $n_b \times n_b$ transition matrix estimated using least squares (with associated Gaussian error-terms ${\mbv \eta}^{(L)}_t $) and $ {\mbv \Sigma}^{(L)} $ is a  $n_z \times n_z$ covariance matrix estimated empirically using the in-sample residuals from the model fit (i.e., this is essentially a two-stage least squares estimation procedure, where $ {\mbv \Phi} $ is first estimated as previously described, followed by $\bM$). See Chapter 7 of Cressie and Wikle (2011) for a comprehensive overview of linear DSTMs. 

The models considered here are evaluated in terms of both out-of-sample prediction accuracy, and the quality of their respective forecast distributions (i.e., uncertainty quantification). In particular, mean squared prediction error (MSPE), defined here as the average squared difference between out-of-sample realizations and forecasts averaged over space and time, is calculated for every model. The forecast distributions are evaluated using the continuous ranked probability score (CRPS). Unlike MSPE, CRPS also considers the quality of a given model's uncertainty quantification by comparing the forecast distribution with the observed values \citep[e.g., see][]{gneiting2014probabilistic}. The classic CRPS formulation for Gaussian data \citep[i.e.,][]{gneiting2005calibrated} is used for the soil moisture application, while the deep Lorenz-96 simulation uses the closed form expression for CRPS with log-Gaussian data from \cite{baran2015log}. Regarding the soil moisture application, as previously noted, standard forecasting methodologies for difficult long-lead problems are often compared to simpler forecast methods such as climatological or linear forecasts.  The so-called {\it skill score} (SS) is an evaluation metric that compares a forecasting method to some reference forecast \citep[e.g.,][]{wilks2001skill}. Assuming one wants to compare a {\it reference} forecast to some {\it model} forecast in terms of MSPE, SS is defined as follows:
\begin{equation}
\text{SS}=1-\frac{\text{MSPE}(Model)}{\text{MSPE}(Reference)}.
\label{SS_B_DEESN}
\end{equation} 
In our application, SS is calculated for each location in the soil moisture application by calculating MSPE across time. Thus, by computing a SS for each spatial location we can obtain a spatial field that shows where the new model improves upon a particular reference model.

\subsection{Simulation Study: Deep Lorenz-96 Model}
\label{deepLorenz96}
The deterministic model of \citet[][]{lorenz1996predictability}, often referred to as the ``Lorenz-96 model,'' is frequently used as a simulation model in the atmospheric science and dynamical systems literature because it incorporates the quadratic nonlinear interactions of the famous \citet[][]{lorenz1963deterministic} ``butterfly'' model in a one-dimensional spatial setting.  A multiscale extension of this model \citep[see][]{wilks2005effects} has gained popularity in the atmospheric science and applied mathematics literature for its ability to represent realistic nonlinear interactions between small and large scale variables \citep[][]{grooms2015framework}. Moreover, the Lorenz-96 model operates on multiple scales in both space and time, consisting of locations with slowly varying large-scale behavior and locations with fast varying small-scale behavior. Thus, the multiscale Lorenz-96 model represents a very relevant example for the multiscale deep EESN methodology developed here.  To make this simulation model even more realistic, we extend it  by adding an additional data stage to allow for non-Gaussian data types. We will refer to this simulation model as the ``deep Lorenz-96 model.'' To our knowledge, this deep Lorenz-96 model is novel.

\begin{figure}[H]
  \centering
\includegraphics[width=9.5cm,height=9.5cm]{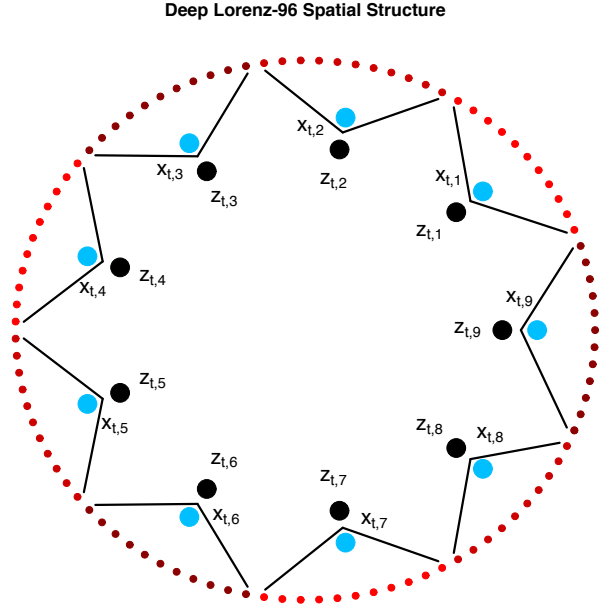}
\caption{Description of the spatial structure for the deep Lorenz-96 model with 9 large-scale locations that each have 11 associated small-scale locations, which gives 99 small-scale locations. For the sake of visualization we have used a smaller number of large and small scale locations here than in the analyzed deep Lorenz-96 simulated data example. The various red-shaded small circles represent small-scale locations (i.e., $y_{j,k}$ in (\ref{deepLorenz})), while the large blue and black circles denote the large-scale locations  $x_k$ and $z_k$, respectively, from the process model defined in (\ref{deepLorenz}).}
\label{fig:Figure_1}
\end{figure}

The spatial structure of the Lorenz-96 system (as shown in Figure \ref{fig:Figure_1}) is a one-dimensional circular structure (i.e., periodic boundary conditions) where each large-scale location is evenly spaced. Furthermore, each large-scale location is associated with $J$ small scale locations (see Figure \ref{fig:Figure_1}). Using the  process model parameterization from \cite{chorin2015discrete} and the inclusion of a data model, our deep Lorenz-96 model is defined as follows for time period $t$:
\begin{eqnarray}
\mbox{Data Model:} & \; & z_{t,k}\sim \text{LogGaussian}( \frac{ | x_{t,k}|}{c},\sigma^2_\eta),  \nonumber \\
\mbox{Process Model:} & \; &  \frac{dx_{k}}{dt} = x_{k-1}(x_{k+1} - x_{k-2}) - x_{k} + F+ \frac{h_x}{J} \sum_{j} y_{j,k},   \label{eq:L96} \nonumber \\
& \; &   \frac{dy_{j,k}}{dt} =\frac{1}{\epsilon_L}[ y_{j+1,k}(y_{j-1,k}-y_{j+2,k})-y_{j,k}  + h_y x_{k} ] \label{deepLorenz}, 
\end{eqnarray}
where $c,F,\epsilon_L,h_x,$ and $h_y$ are user defined parameters, $z_{t,k}$ and $x_{t,k}$ correspond to large-scale locations, $y_{j,k}$ corresponds to a small-scale location for $j=1,\dots, J$ and $k=1,\dots, K$, and $| \cdot |$ denotes the absolute value. The parameter $F$ corresponds to a forcing term and helps determine the amount of nonlinearity in the model.  In this setting, $\epsilon_L$ is a ``time separation parameter'' such that smaller values lead to a faster-varying temporal-scale for the small-scale locations. The contribution of the small-scale locations to the large-scale locations (and vice versa) is determine by $h_x$ and $h_y$, respectively. Finally, the scaling parameter $c$ helps ensure the log-Gaussian data model does not produce unrealistically large realizations (we let  $c=2$ here). 

\begin{figure}[H]
  \centering
\includegraphics[width=15cm,height=11cm]{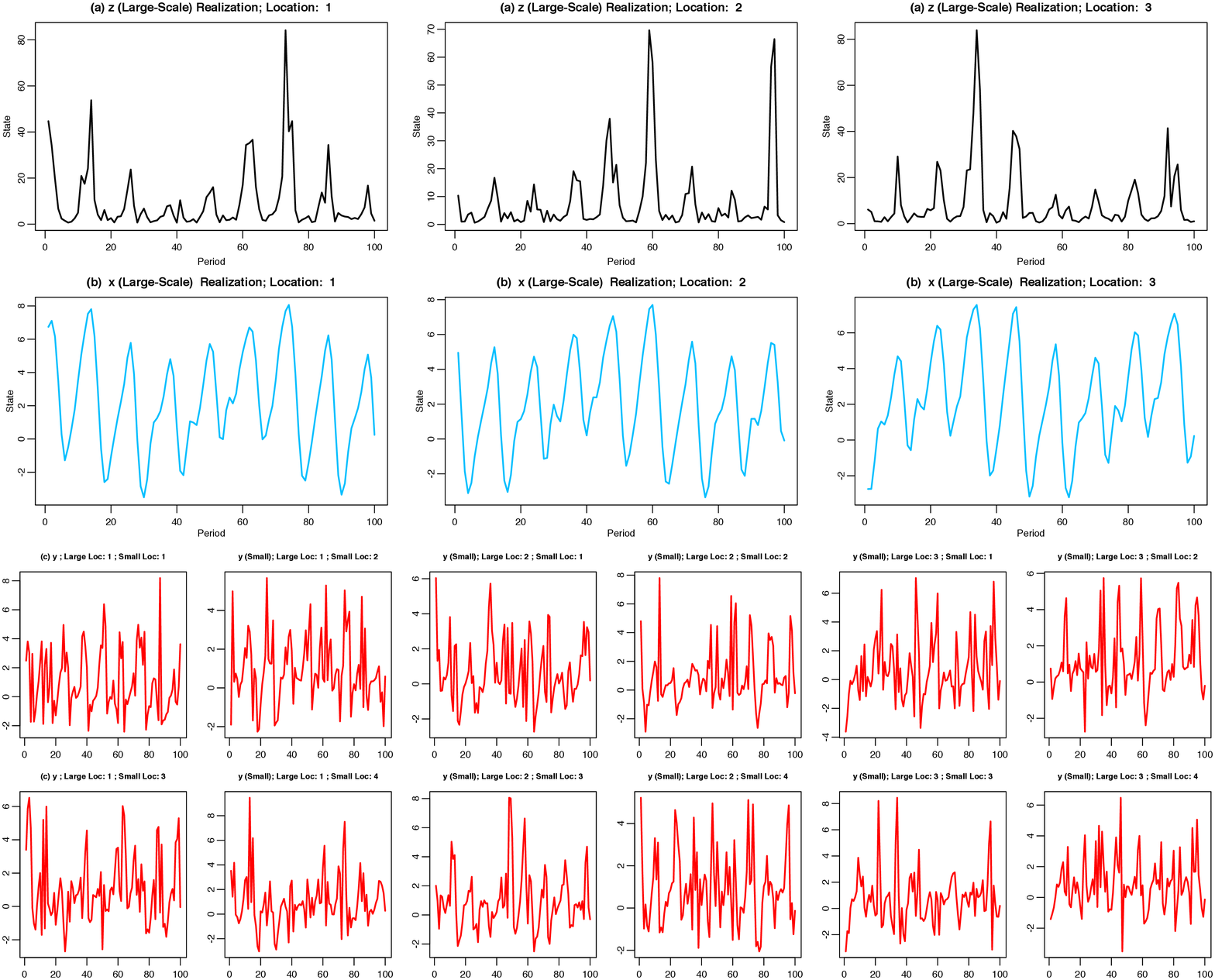}
\caption{A simulation from the deep Lorenz-96 model defined in (\ref{deepLorenz}) for 100 time periods. (a) Realizations from the data model in (\ref{deepLorenz}) for three locations where these locations are transformations of the slowly varying large-scale locations shown in the second row (i.e., (b)). (b) Realizations for three large-scale locations from the process model defined in (\ref{deepLorenz}). (c) Realizations corresponding to small-scale locations for each of the three large-scale locations displayed in the second row. Note, each column displays four (of the twenty) slow varying small-scale locations associated with a particular large-scale location.}  
\label{fig:Figure_2}
\end{figure}

Since the methodology presented here concerns multiscale spatio-temporal modeling, the parameters for the deep Lorenz-96 model are selected to emphasize the level of multiscale behavior in the process (as can be seen from the simulated deep Lorenz-96 data shown in Figure \ref{fig:Figure_2}). In particular we use the following settings:  $h_x=-1.90, h_y=1$, and $\eta_L=.045$, while we follow \cite{chorin2015discrete} in setting $F=10,K=18,$ and $J=20$. The variance term in (\ref{deepLorenz}) is set such that $\sigma^2_\eta=.25$. An Euler solver is used to numerically solve the Lorenz-96 equations in (\ref{eq:L96}) using a time step of $\delta=.10$. After a burn-in period, 510 periods of simulated data from the deep Lorenz-96 model are retained, with the final 75 periods held out and treated as out-of-sample realizations.

Only the large-scale process $z_{t,k}$ is treated as observed here, thus both the large and small scale Lorenz-96 variables (i.e., $x_{k}$ and $y_{j,k}$) are considered unobserved. Therefore, unlike the soil moisture application, both the input and output of the model are the same process (i.e., $z_{t,k}$). The input and output are separated by three periods here (i.e., the lead time is set to three periods), in order to make the problem slightly more difficult. The previously discussed embedding lag (i.e., $\tau$ in (\ref{eq:QESNembed})) is set to the lead time (three periods) for the deep Lorenz-96 D-EESN implementations. The number of embedding lags (i.e., $m$ in (\ref{eq:QESNembed})) is selected with the GA (using cross-validation) to be three.   Finishing the specification of the data models in (\ref{eq:D-EESNdata}) and (\ref{eq:BD-EESNdata}), ${\mbv \Phi}$ is defined as ${\mbv \Phi}\equiv \bI$, where $\bI$ is a $18 \times 18$ identity matrix, a log-Gaussian distribution is used for the unspecified distribution in (\ref{eq:BD-EESNdata}), and the covariance matrix is set such that ${\mbv \Sigma}_z= \sigma^2_z \bI$, where $\sigma^2_z = \sigma^2_\eta$. 

\begin{table}[H]
\centering
\begin{tabular}{|c|cc|} 
\hline
Model &  MSPE  &  CRPS  \\
\hline
Q-EESN &  82.64 &   3876.45  \\ 
BQ-EESN&   81.51 & 3875.61  \\ 
D-EESN (7 L) & {\bf 76.00} &  {\bf 3872.72}  \\ 
BD-EESN (7 L) & 79.08 & 3873.87     \\ 
Lin. DSTM & 100.12 &  4557.78     \\
\hline
\end{tabular}
\caption{Results for the deep Lorenz-96 simulation study in terms of mean squared prediction error (MSPE) and continuous ranked probability score (CRPS). {\it Q-EESN} refers to the quadratic EESN; {\it BQ-EESN } denotes the Bayesian version of the  Q-EESN model; {\it D-EESN (7 L)} is the D-EESN model with seven layers, and {\it BD-EESN (7 L)} denotes the BD-EESN model with seven layers. Note, smaller is better for both MSPE and CRPS.}
\label{tab:Table_3}
\end{table}

The out-of-sample validation metrics in Table \ref{tab:Table_3} show the seven-layer D-EESN models to be the best forecast model for the deep Lorenz-96 data. In particular, both of the seven-layer D-EESN models perform better in terms of MSPE and CRPS than the Q-EESN or linear model. Although not shown here, D-EESN models with 2-6 layers all performed better than the Q-EESN model and worse than the seven-layer D-EESN model in terms of MSPE, with the MSPE monotonically decreasing as layers were added. We found that after seven-layers any gain in forecast accuracy was very minimal in comparison to the extra computation required to keep adding layers.

\begin{figure}[H]
  \centering
\includegraphics[width=16cm,height=8.5cm]{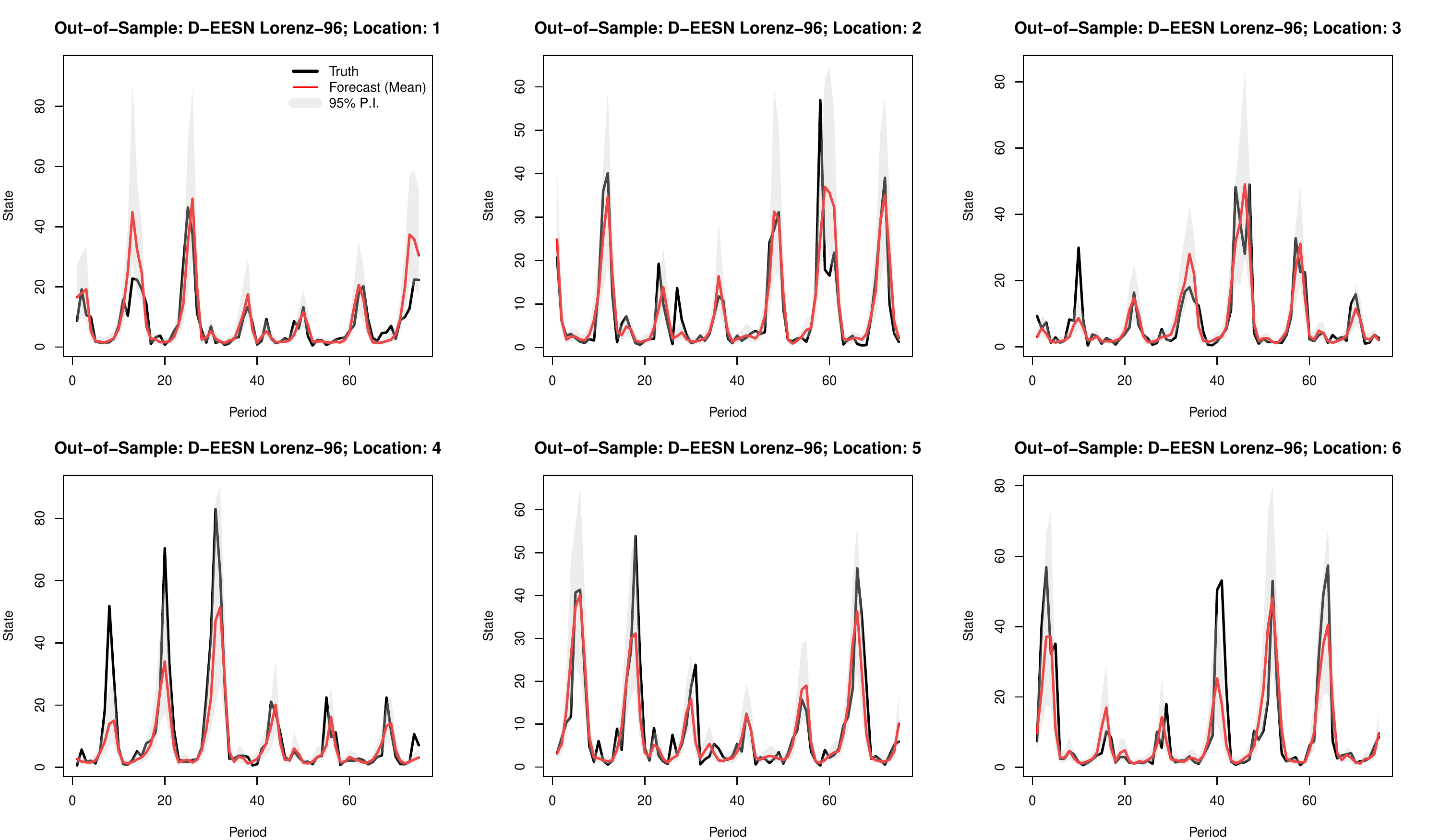}
\caption{Out-of-sample forecast summary plots for 6 of the 18 large-scale locations from the deep Lorenz-96 simulation study using the seven layer D-EESN model. The black line denotes the true simulated value in each plot (i.e., $ z_{t,k}$ in (\ref{deepLorenz})), while the red line denotes the forecasted mean. In a given plot, the shaded gray area represents the 95\% point-wise prediction intervals (P.I.s) over all ensembles.}
\label{fig:Figure_3}
\end{figure}

The Bayesian version of the seven-layer D-EESN model produces similar MSPE and CRPS values to the non-Bayesian version. Although one would expect the Bayesian model to perform similar in terms of MSPE to the non-Bayesian version, there is potential for the Bayesian version to improve in terms of uncertainty quantification (as shown below for the soil moisture application below). This model's ability to quantify uncertainty is illustrated in Figure \ref{fig:Figure_3}, which shows forecast summaries for the first six locations with the seven-layer D-EESN model. Despite the clear nonlinearity in the process, the model correctly forecasts much of the overall quasi-cyclic nature and intensity of the process, while producing uncertainty metrics that cover many of the true values.

\subsection{Midwest Soil Moisture Long-Lead Forecasting Application }\label{sec:SoilMoistureEx}
The previously mentioned soil moisture data comes from the Climate Prediction Center's (CPC) high resolution monthly global soil moisture data set  (\cite{fan2004climate}, \verb+https://iridl.ldeo.columbia.edu/SOURCES/.NOAA/.NCEP/.CPC/.GM+ \\ \verb+SM/.w/+). As described in \cite{smith2008improvements}, the driving inputs used to create this derived data consist of global precipitation and temperature data along with an accompanying land model. Spatially, the data domain covers $35.75^{\circ}$N - $48.75^{\circ}$N latitude and $101.75^{\circ}$W - $80.25^{\circ}$W longitude at a resolution of $0.5^{\circ} \times 0.5^{\circ}$. The monthly data set begins in January 1948 and goes through December 2017. While the model is trained on monthly data from January 1948 through November 2011, only the May forecasted values from the out-of-sample period covering 2012-2017 are used to evaluate the model, given the importance of May soil moisture for planting corn in the U.S. corn belt.  We follow the common practice in the atmospheric science literature of converting data into anomalies by subtracting the respective in-sample monthly means from the data.

Dimension reduction for the soil moisture data is carried out using empirical orthogonal functions (EOFs; i.e., or spatial-temporal principal component analysis) \citep[e.g., see][Chapter 5]{CandW2011}. Therefore, ${\mbv \Phi}$ is obtained using the first 15 EOFs, which account for $80\%$ of the variation in the soil moisture data (note, the model was not sensitive to small variations in this choice). The data model spatial covariance matrix ${\mbv \Sigma}_z$ is calculated using the following formulation from \citet[][Equation 7.6]{CandW2011}: ${\mbv \Sigma}_z=\sum\limits_{\ell_z=n_b+1}^{n_z} \tilde{\lambda}_{\ell_z} {\mbv \Phi}_{\ell_z} {\mbv \Phi}_{\ell_z}' + \tilde{c} \ \bI$, where $\tilde{\lambda}_{\ell_z}$ and ${\mbv \Phi}_{\ell_z}$ are the remaining eigenvalues and eigenvectors, respectively, that are not used in the decomposition of $\bZ_t$ and $\tilde{c} $ is a constant (set to $0.01$). Since the data are converted into anomalies, a Gaussian distribution is used for the distribution in (\ref{eq:BD-EESNdata}).

The monthly SST data comes from the extended reconstruction sea surface temperature (ERSST) data set (\cite{smith2008improvements}, \verb+http://iridl.ldeo.columbia.edu/+ \\ \verb+SOURCES/.NOAA/.NCDC/.ERSST+) and cover the same temporal period as the soil moisture data. The spatial domain for the SST data is given by $29^{\circ}$S - $29^{\circ}$N latitude and $124^{\circ}$E - $70^{\circ}$W longitude with a resolution of $2^{\circ} \times 2^{\circ} $, and covers much of the mid-Pacific ocean. Dimension reduction is once again carried out using EOFs by retaining the first 5 EOFs of the SST data set, which account for almost $72\%$ of the variation in the data (the model was also not overly sensitive to this choice). The embedding lag for the input is set to the lead time of six periods and the number of embedding lags is selected to be three using cross-validation with the GA. Convergence for the MCMC Gibbs algorithm sampler was further assessed using the Gelman-Rubin diagnostic \citep[][]{gelman1992inference} with four chains, which did not suggest any lack of convergence.

\begin{table}[H]
\centering
\begin{tabular}{|c|ccc|} 
\hline
Model &  MSPE  &  CRPS & \% SS$>$0  \\
\hline
Q-EESN & 3507.84  & 239.16  & 50.60\%   \\ 
BQ-EESN  & 3463.28  & 196.59  & 55.20\%   \\ 
D-EESN (2 L) &  3303.64 & 229.96 & 60.00\%\\ 
BD-EESN (2 L) & {\bf 3296.39}  & 190.45 & 58.24\% \\ 
D-EESN (3 L) &  3307.51 & 224.68 & 60.32  \\ 
BD-EESN (3 L) & 3299.04  & {\bf 189.80}  & {\bf 63.54\%}  \\ 
Lin. DSTM & 3509.63 & 198.09 & 50.00\%   \\
Climatological & 3642.54 &  - & -  \\
\hline
\end{tabular}
\caption{Validation results for the long-lead soil moisture forecasting application using mean squared prediction error (MSPE), continuous ranked probability score (CRPS), and percentage of skill score values greater than zero (i.e., \% SS$>$0 ). The prefix ``B'' denotes a Bayesian version of the model and and ``L'' denotes the number of layers in a given model.}
\label{tab:Table_4}
\end{table}

Table \ref{tab:Table_4} shows the performance of various models for out-of-sample long-lead forecasting of May soil moisture.  Both the Bayesian and non-Bayesian D-EESN models perform the best in terms of MSPE. The D-EESN models also represent the largest improvement over the climatological forecast. Further, both D-EESN models also outperform the single-layered Q-EESN model, suggesting the soil moisture application benefits from a deep framework.  Notably, the two and three layer models appear to have similar MSPE values, indicating that two layers is likely sufficient.  While the Bayesian and non-Bayesian models  perform similarly in terms of forecast accuracy, the Bayesian D-EESN models perform much better in terms of CRPS than the non-Bayesian versions. In particular, the Bayesian version produces CRPS values that are almost $16\%$ lower than the non-Bayesian version. Absent the improvement allowed by the Bayesian implementation, the D-EESN would be much less able to quantify the uncertainty in the long-lead soil moisture forecasts.

\begin{figure}[H]
  \centering
\includegraphics[width=11cm,height=12cm]{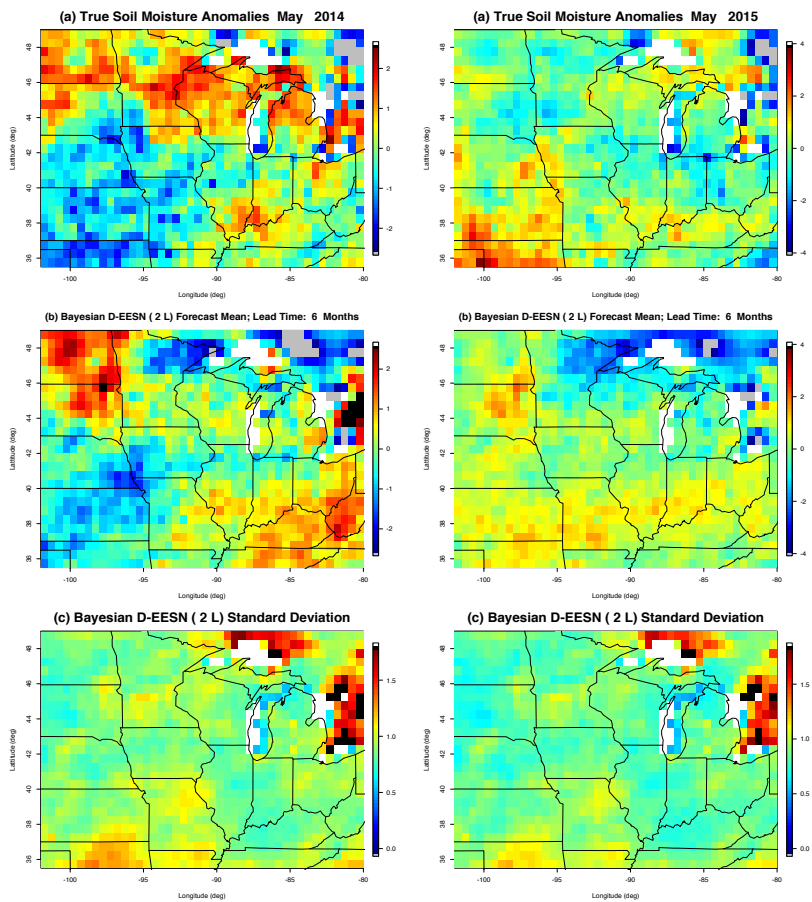}
\caption{Posterior summaries for the soil moisture application in May 2014 and 2015 using the two-layer BD-EESN model. (a) True soil moisture moisture values for each spatial location. (b) Posterior predictive mean values for each spatial location. (c) Posterior predictive standard deviations for each spatial location. Note, for the sake of visualization, each plot has been standardized by their respective means and standard deviations. Also, note that the scale bars are different for the 2014 and 2015 anomalies to improve forecast contrast. We have removed extreme outliers for the sake of visualization (indicated by grey and black grid squares).}  
\label{fig:Figure_4}
\end{figure}

\begin{figure}[H]
  \centering
\includegraphics[width=11cm,height=12cm]{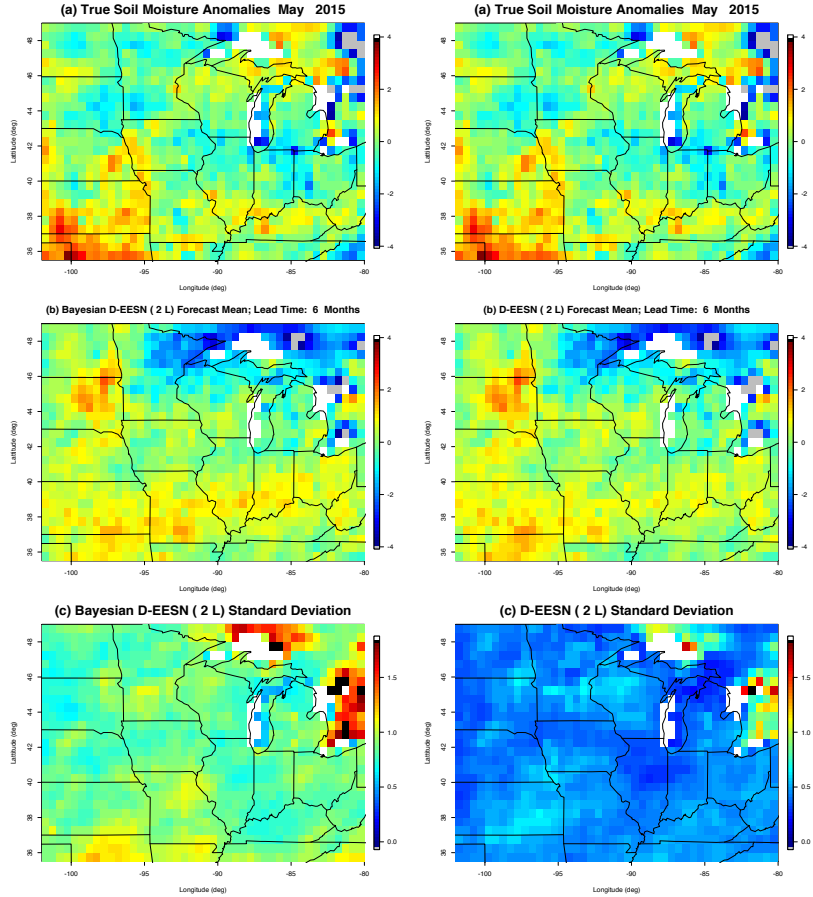}
\caption{May 2015 forecast summaries for both the Bayesian and non-Bayesian two-layer D-EESN model. (a) True soil moisture moisture values for each spatial location. (b) Forecasted mean values with a given forecasting method for each spatial location. (c) Forecast standard deviations with a given forecasting method for each spatial location. Note, for the sake of visualization each plot has been standardized by their respective means and standard deviations. We have removed extreme outliers for the sake of visualization (indicated by grey and black grid squares).}  
\label{fig:Figure_5}
\end{figure}

Figure  \ref{fig:Figure_4} shows the posterior predictive means and standard deviations for 2014 and 2015 over the prediction domain as given by the two-layer BD-EESN model. The model appears to mostly pick up the correct pattern of  the soil moisture anomaly signal in the western and mid-central part of the spatial domain for both years, while struggling more with the upper midwest states in 2014. Regarding the critical corn belt, the model suggests an overall large amount of uncertainty in Missouri, especially in 2014, while Iowa appears to have a moderate to low amount of uncertainty for both years. As previously noted, the BD-EESN produces considerably better uncertainty metrics for the soil moisture data than the non-Bayesian D-EESN. The relative difference between these two predictive uncertainties can be seen in Figure \ref{fig:Figure_5}, where both versions of the two-layer D-EESN model are plotted for 2015. Although the forecast means are very similar in Figure \ref{fig:Figure_5}, the non-Bayesian D-EESN produces much smaller standard deviations across the spatial domain compared to the Bayesian version. Considering the inherent difficulty in predicting soil moisture six months into the future, the standard deviations for the non-Bayesian model appear unrealistically low. This point is confirmed by the Bayesian model producing a considerably lower CRPS value than the non-Bayesian version.

\begin{figure}[H]
  \centering
\includegraphics[width=8.5cm,height=8.5cm]{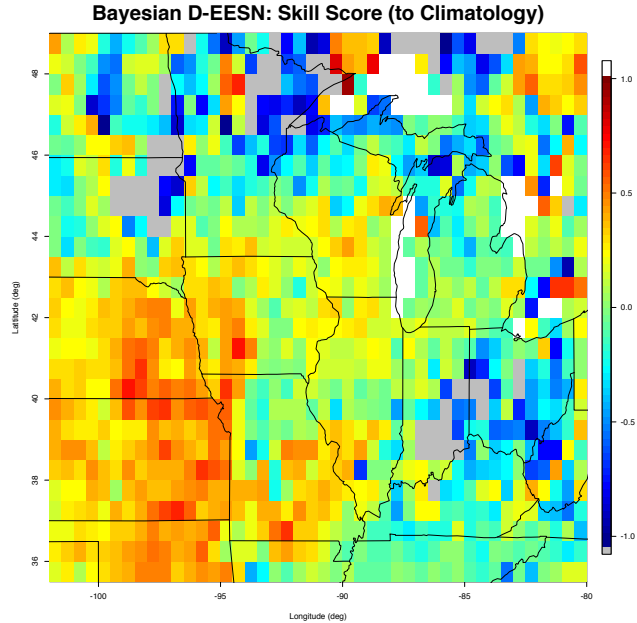}
\caption{Skill score (SS) plot for the soil moisture application with the Bayesian two-layer D-EESN model. SS is calculated using (\ref{SS_B_DEESN}) from above, where the {\it reference model} used here is a climatological forecast. For each forecasting method, the MSPE is calculated by averaging over the out-of-sample periods. Values of SS greater than zero indicate an improvement of the two-layer BD-EESN model over the climatological forecast, while values less than zero indicate locations where the two-layer BD-EESN model performed worse than climatology. We have removed extreme outliers for the sake of visualization (indicated by grey grid squares).}  
\label{fig:Figure_6}
\end{figure}

Next, the previously defined skill score (SS)  in (\ref{SS_B_DEESN}) is shown in Figure \ref{fig:Figure_6} for the two-layer BD-EESN model, where a climatological forecast is used as the reference model. Values of SS greater than zero represent locations where the BD-EESN model improved upon the climatological forecast, while values less than zero indicate locations where the model did worse than the climatological forecast. Overall, the BD-EESN model outperforms the climatological forecast in the central western part of the domain, and performs worse in the east central and northern part of the domain. In particular, the model does worse in these regions by predicting too little soil moisture in the north central part of the domain, relative to the truth, and over predicts the amount of soil moisture in the east central part of the domain. Critically, across much of the agriculturally important corn belt, including much of Iowa, the BD-EESN model improves upon the climatological forecast.

\section{Discussion}

\begin{figure}[H]
  \centering
\includegraphics[width=9cm,height=7.5cm]{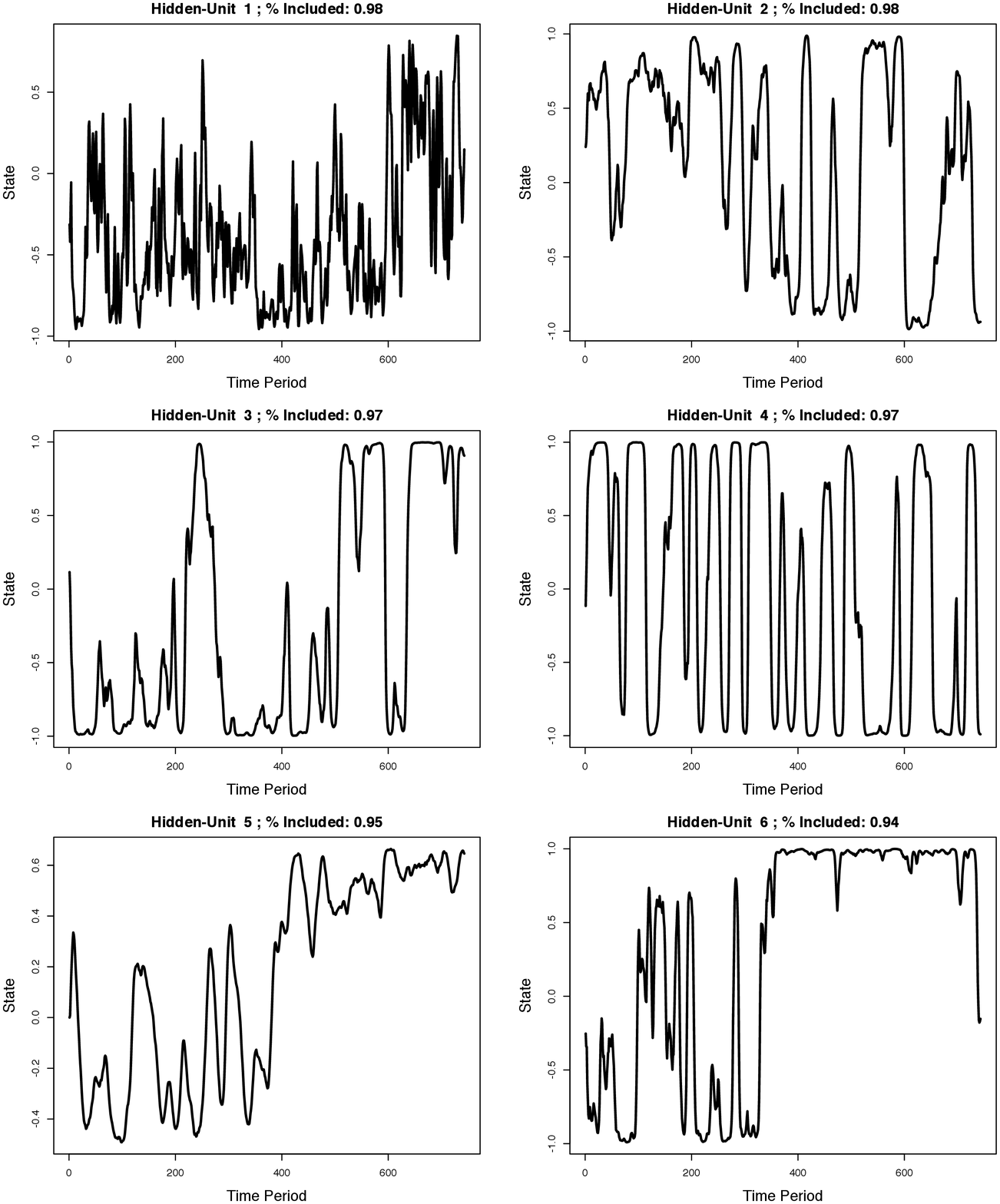}
\caption{Time series of the six most chosen hidden unit predictors used to predict the first soil moisture EOF coefficient ($\alpha_t(1)$) as chosen via SSVS in the BD-EESN model.  The percentage of MCMC iterations that include each predictor is given shown above each series.}  
\label{fig:Figure7}
\end{figure}

The models presented here are essentially just regularized spatio-temporal regression models where the inputs (predictors) are stochastically and dynamically transformed.  Indeed, the model presented in (\ref{eq:BD-EESNdata}) and (\ref{eq:BD-EESNoutput}) might be considered a generalized additive model applied to spatio-temporal data, where the inputs are transformed through a telescoping set of ESN models.  But, because of the inherent randomness in the choice of reservoir parameters in the ESNs, we have to do many independent replications of these transformations in order to obtain reproducible results.  This is the nature of stochastic transformations.  Note that these replications of the deep ESN are given equal weight in  (\ref{eq:BD-EESNoutput}), but this need not be the case in general.  It is also important to note that although the spatio-temporal regression model is not strictly dynamic (in the sense of a spatial field evolving in time), the {\it transformations are inherently dynamic} through the ESN recurrence relation. Finally, it is important to note that the multiple levels of transformation allow for different time scales in the predictors -- this is the advantage of the deep architecture.  How do we know that deep structure is giving multi-time scale predictors?  Figure \ref{fig:Figure7} shows the most frequently chosen predictors for the first EOF coefficient in the soil moisture example ($\alpha_t(1)$).  These predictors exhibit different time scales.  Although this shows that different time scales are important for the prediction, it is not so clear how to interpret these predictors given the complex nonlinear transformations that generated them in the deep ESN structure.  This lack of interpretability is a fundamental problem with many deep models in the machine learning context as well, and is an active area of investigation in statistics and machine learning.

Despite the large amount of research into deep models, uncertainty quantification is rarely considered in their application. Given the demonstrated predictive ability of these methods, having the ability to quantify uncertainty with deep models is very powerful and widely applicable. Through the use of an ensemble framework, the deep ESN models for spatio-temporal processes presented here allow for uncertainty quantification. The non-Gaussian Lorenz-96 simulated example showed that the D-EESN improved upon the traditional ESN model, while also producing very robust estimates of the forecast uncertainty. Furthermore, the Bayesian version of the D-EESN model provides a formal framework in which many data types can be considered and multiple levels of uncertainties can be accounted for. The results for the BD-EESN with the soil moisture data illustrated this point by considerably improving upon the uncertainty metrics produced by the D-EESN model.  We were also able to show spatially where the deep models improved upon simpler forecast methods, thus giving model developers and resource managers potentially useful information. 

As discussed above, there are many more possible choices for the dimension reduction function used with the hidden units from the D-EESN model. One potential choice that has been unexplored in the literature is so-called {\it convolution operators}. Deep image classification methods have shown convolution operators to be extremely powerful for slowly learning general (spatial) features \citep[][]{krizhevsky2012imagenet}. Its possible that the deep RNN framework could also benefit from such tools, especially in situations where the input has explicit spatial structure. 

Long-lead spatio-temporal soil moisture forecasting is inherently a challenging problem. It is not uncommon to treat such difficult forecasting problems as categorial instead of continuous. That is, treating the response as categorical by re-labeling continuous values with qualitative values (e.g., below average, average, above average). Unlike the RNN literature, most of the ESN literature has focused on continuous responses. Given the flexibility of the BD-EESN, a categorical model formulation can be developed and is the subject of future research.  Finally, the soil moisture forecasts may also benefit from using other climate indexes or more local variables (such as precipitation) as inputs into the model.

\section*{Acknowledgments}
This work was partially supported by the US National Science Foundation (NSF) and the US Census Bureau under NSF grant SES-1132031, funded through the NSF-Census Research Network (NCRN) program, and NSF award DMS-1811745.

\newpage
\bibliography{reference}
\newpage

\end{document}